\documentclass[a4paper,fleqn]{cas-dc}

\usepackage{balance}
\usepackage[numbers]{natbib}
\usepackage{amsmath,amssymb,amsfonts}

\usepackage{multirow,subfigure}
\usepackage{booktabs,bm}
\usepackage{algorithm}
\usepackage{algorithmic}

\usepackage{amssymb}
\def\method{COOL}

\begin{document}
\let\WriteBookmarks\relax
\def\floatpagepagefraction{1}
\def\textpagefraction{.001}

\shorttitle{COOL: A Conjoint Perspective on Spatio-Temporal Graph Neural Network for Traffic Forecasting}

\shortauthors{Ju and Zhao et al.}

\title [mode = title]{COOL: A Conjoint Perspective on Spatio-Temporal Graph Neural Network for Traffic Forecasting}                      



%

\author[1]{Wei Ju}[orcid=0000-0001-9657-951X]
\fnmark[1]

\author[1]{Yusheng Zhao}
\fnmark[1]

\author[1]{Yifang Qin}

\author[2]{Siyu Yi}

\author[1]{Jingyang Yuan}

\author[3]{Zhiping Xiao}
\author[3]{Xiao Luo}
\cormark[1]
\author[4,5]{Xiting Yan}
\author[1]{Ming Zhang}
\cormark[1]


\affiliation[1]{organization={School of Computer Science, National Key Laboratory for Multimedia Information Processing, Peking University},
    city={Beijing},
    postcode={100871}, 
    country={China}}

\affiliation[2]{organization={School of Statistics and Data Science, Nankai University},
    city={Tianjin},
    postcode={300071}, 
    country={China}} 
    
\affiliation[3]{organization={Department of Computer Science, University of California},
    city={Los Angeles},
    postcode={90095}, 
    country={USA}}

\affiliation[4]{organization={Section of Pulmonary, Critical Care and Sleep Medicine, Yale School of Medicine},
    city={New Haven, CT},
    postcode={06520}, 
    country={USA}}

\affiliation[5]{organization={Department of Biostatistics, Yale School of Public Health},
    city={New Haven, CT},
    postcode={06520}, 
    country={USA}}

\cortext[cor1]{Corresponding authors.}

\fntext[fn1]{Equal contribution with order determined by flipping a coin.}


\begin{abstract}
This paper investigates traffic forecasting, which attempts to forecast the future state of traffic based on historical situations. This problem has received ever-increasing attention in various scenarios and facilitated the development of numerous downstream applications such as urban planning and transportation management. However, the efficacy of existing methods remains sub-optimal due to their tendency to model temporal and spatial relationships independently, thereby inadequately accounting for complex high-order interactions of both worlds. Moreover, the diversity of transitional patterns in traffic forecasting makes them challenging to capture for existing approaches, warranting a deeper exploration of their diversity. Toward this end, this paper proposes \underline{Co}njoint Spati\underline{o}-Tempora\underline{l} graph neural network (abbreviated as \method{}), which models heterogeneous graphs from prior and posterior information to conjointly capture high-order spatio-temporal relationships. On the one hand, heterogeneous graphs connecting sequential observation are constructed to extract composite spatio-temporal relationships via prior message passing. On the other hand, we model dynamic relationships using constructed affinity and penalty graphs, which guide posterior message passing to incorporate complementary semantic information into node representations. Moreover, to capture diverse transitional properties to enhance traffic forecasting, we propose a conjoint self-attention decoder that models diverse temporal patterns from both multi-rank and multi-scale views. Experimental results on four popular benchmark datasets demonstrate that our proposed \method{} provides state-of-the-art performance compared with the competitive baselines.
\end{abstract}



\begin{keywords}
Traffic Flow Prediction \sep Graph Neural Network \sep Spatio-temporal Analysis
\end{keywords}

\maketitle



\section{Introduction}

Spatio-temporal forecasting~\citep{wu2021autocts,li2024survey} has emerged as a prominent research area due to its relevance in numerous downstream applications. From urban planning and environmental management to logistics optimization and beyond, accurate predictions of how entities evolve over time and space are essential. One particularly critical real-world problem in this domain is traffic flow forecasting~\citep{ji2022stden,xie2020urban,xu2023generic,zhao2023dynamic}, which aims to forecast future traffic based on historical situations. It involves predicting various aspects of traffic dynamics, including traffic volume, speed, and congestion patterns, across different locations and time intervals. The applications of traffic flow prediction are far-reaching, impacting intelligent transportation systems~\citep{ji2022stden}, traffic management~\citep{ouallane2022fusion}, route planning~\citep{luo2022diversified}, and ultimately contributing to reduced congestion, improved transportation efficiency, and enhanced urban livability.

Recently, a range of algorithms for effective traffic flow prediction have been proposed, broadly categorized as physics-based and learning-based. For the former, physics-based methods typically rely on differential equations to formally describe traffic systems~\citep{ni2015traffic, di2021survey}. They often exhibit outstanding performance in simulated environments, supported by rigorous theoretical foundations. However, these methods often struggle to adapt to the complexities of real-world scenarios due to their demanding model assumptions~\citep{mo2021physics}. Conversely, for the latter, learning-based approaches are widely adopted for their ability to optimize machine learning models using historical observations, making them a popular choice for predicting future trends. Early research endeavors attempt to tackle this challenge with traditional models such as Autoregressive Integrated Moving Average~\citep{williams2003modeling} and Support Vector Machines~\citep{sun2014traffic}. However, their modeling capacity is insufficient to fit large-scale and complex data. In recent years, deep learning-based approaches have gained considerable attention due to their leverage of the powerful representation learning capabilities of deep neural networks, resulting in significant improvements.  On the one hand, they employ Recurrent Neural Networks~\citep{li2017diffusion} or Temporal Convolutional Networks~\citep{wu2019graph} to capture temporal dependencies in the traffic data. TE-TCN~\citep{ren2023transformer} proposes Transformer-enhanced temporal convolutional networks to capture long- and short-term periodic dependencies. On the other hand, Graph Neural Networks (GNNs) are utilized to extract structured spatial relationships from road networks. DyHSL~\citep{zhao2023dynamic} leverages hypergraph structural information for non-pairwise spatial relationships. By harnessing the strengths of both worlds, current algorithms can effectively capture temporal and spatial information, facilitating accurate traffic flow predictions~\citep{xu2023generic,guo2019attention,song2020spatial,fang2021spatial,liu2023spatial}. For example, GDGCN~\citep{xu2023generic} develops a novel temporal graph convolutional block for flexible temporal relations, and a dynamic graph constructor to model both time-specific spatial dependencies and changing temporal interactions. DS-TGCN~\citep{liu2023spatial} incorporates spatial–temporal similarity feature and convolution with an attention mechanism to effectively extract complex spatial–temporal relationships.

\begin{figure}[t]
\centering
\setlength{\abovecaptionskip}{0cm}
\setlength{\belowcaptionskip}{-0.5cm}
\subfigure[Different variances]{
\label{fig:mot:a}
\includegraphics[width=0.231\textwidth]{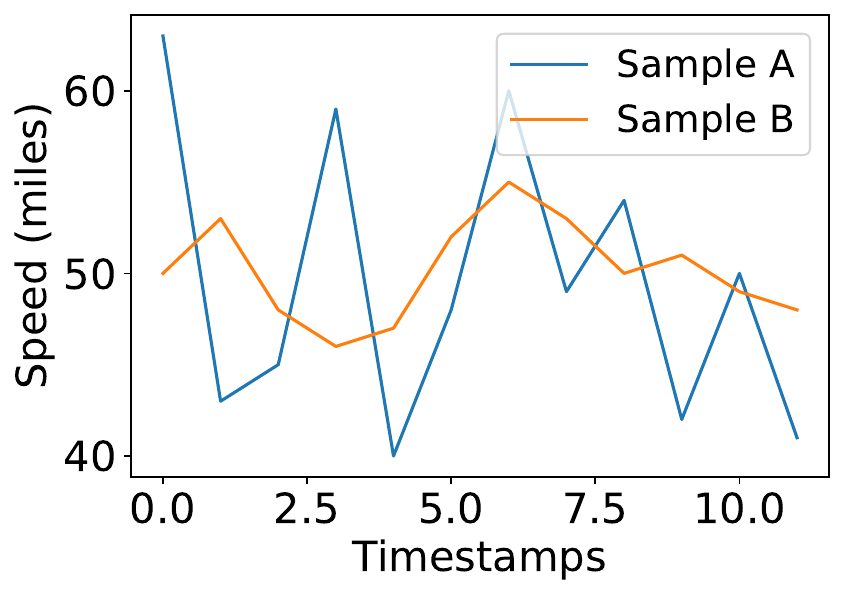}}
\subfigure[Different periodicity]{
\label{fig:mot:b}
\includegraphics[width=0.231\textwidth]{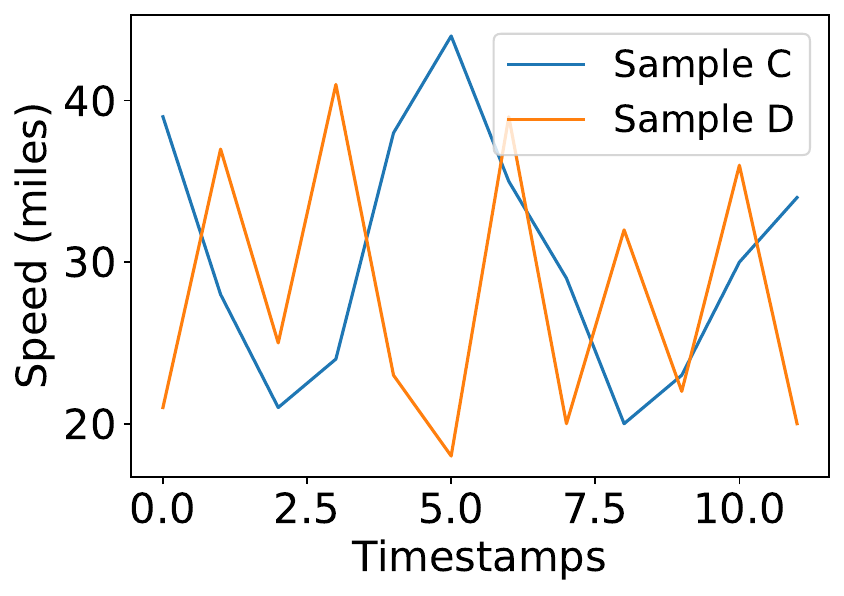}}
\caption{Transitional patterns in traffic networks could be diverse, varying in terms of variances (a) and periodicity (b).}
\label{fig:mot}
\vspace{-0.1cm}
\end{figure}

Nevertheless, despite the impressive performance achieved by existing traffic flow forecasting techniques~\citep{pholsena2020mode,song2020spatial,wang2021metro,yang2022inductive,saenz2023nation}, they still suffer from two critical flaws: 
(1) \textit{Fail to effectively capture composite spatio-temporal relationships.} For example, consider the failure to adequately model the interplay between traffic congestion and local weather conditions, such as rain or snow, which can significantly impact traffic flow dynamics. Existing spatio-temporal algorithms typically combine GNNs and RNNs by fusing the corresponding representations in traffic networks~\citep{zheng2020gman,wu2020connecting,li2023dynamic}. Unfortunately, this naive combination separates the mining of spatial and temporal correlations. In detail, these methods~\citep{qu2019daily,luo2019spatiotemporal} cannot acquire various temporal information while extracting spatial messages, thus missing high-order composite relationships. Worse still, they usually extract spatial correlations from road networks, which neglects dynamic semantic correlations in traffic systems, resulting in inferior performance for traffic prediction. 
(2) \textit{Failing to sufficiently capture diverse transitional patterns.} Due to diverse traffic requirements, different locations or times could exhibit various transitional patterns. For example, existing models may have difficulty accurately predicting transitions between regular weekday traffic flow and the highly variable traffic patterns during holidays or special events like concerts or sports games. Furthermore, they might not effectively capture the transition patterns between workdays and weekends, which often exhibit distinct traffic dynamics due to changes in commuter behavior. As shown in Figure \ref{fig:mot}, locations could show different periodic patterns due to daily or weekly routines~\citep{guo2019attention}. However, existing methods mostly fail to model complex temporal dependencies effectively using standard sequential models, hindering them from making accurate traffic low predictions.

In this paper, we present a novel approach named \method{} for effective traffic prediction. At a high level, \method{} conjointly explores high-order spatio-temporal relationships from both prior and posterior information. On the one hand, we introduce prior information into constructed heterogeneous graphs connected by spatial and temporal connections. On the other hand, we model dynamic relationships using both constructed affinity and penalty graphs, then a posterior message passing layer is developed to incorporate both similarities and dissimilarities into sequential node representations. In addition, to capture diverse transitional properties to enhance traffic forecasting, we develop a conjoint self-attention decoder that aggregates sequential representations by modeling diverse temporal patterns from both multi-rank and multi-scale views. In particular, we not only utilize transformation matrices of different sizes to provide multi-rank attention matrices to model diverse intrinsic patterns, but also involve multi-scale pooling to generate subsequence representations for capturing various periodic patterns. Eventually, we combine obtained global representations adaptively to generate traffic predictions. 

Compared to the current state-of-the-art baseline method STAEformer~\citep{liu2023spatio}, which primarily employs different Transformer layers to sequentially model temporal and spatial information, independently modeling them for two inherently coupled aspects often leads to sub-optimal performance. In contrast, our proposed COOL naturally couples temporal and spatial information by constructing a heterogeneous graph. Through both prior and posterior message passing, COOL further captures high-order spatio-temporal dependencies. Additionally, STAEformer does not consider the rich temporal patterns present in real-world scenarios of traffic flow prediction, while our COOL leverages the self-attention mechanism~\citep{vaswani2017attention} to model these patterns and capture dynamics more realistically.

The main contributions of this paper are as follows:

\begin{itemize}
    \item We propose a novel spatio-temporal graph convolutional network model \method{}, which conjointly explores high-order spatio-temporal relationships using both prior and posterior information.
    \item To capture various long-term transitional patterns, \method{} introduces a conjoint self-attention decoder that aggregates sequential representations using both multi-rank and multi-scale attention branches.
    \item Extensive experiments on four benchmark datasets achieve promising results and outperform the competitive baseline methods by a large margin, which validates the effectiveness of our method.
\end{itemize}


\section{Related Work}
\label{sec::related}


\subsection{Graph Neural Networks}
GNNs have gained significant popularity in recent years due to their effectiveness in modeling structured data with complex relationships~\citep{ju2023comprehensive,holzinger2021towards,zhou2022identifying,luo2023hope,huang2023attentive,passos2023multimodal,ju2024survey}. The fundamental idea behind GNNs is to learn representations of nodes in a graph by aggregating information from their neighbors, thereby capturing the inherent graph structure~\citep{gilmer2017neural}. GNNs have been extensively applied to various downstream tasks, including node classification~\citep{kipf2017semi,fang2022polarized,luo2023toward}, graph classification~\citep{ying2018hierarchical,lee2019self,luo2023rignn,luo2023towards,yuan2023learning,ju2023tgnn}, and graph clustering~\citep{yue2022survey,ju2023glcc,yi2023redundancy}, where they have demonstrated impressive performance. In the context of spatio-temporal analysis, GNNs have been extensively applied. Several previous works in spatio-temporal analysis have utilized GNNs to model various phenomena~\citep{wu2019graph,song2020spatial,meng2021cross,shao2022pre,zhao2023dynamic}. For example, Graph WaveNet~\citep{wu2019graph} captures spatial-temporal dependencies using an adaptive dependency matrix learned through node embeddings, and handles long sequences efficiently with stacked dilated 1D convolution components. STSGCN~\citep{song2020spatial} efficiently captures complex localized spatial-temporal correlations using a designed synchronous modeling mechanism and accounts for heterogeneities in localized spatial-temporal graphs through multiple time period modules. CNFGNN~\citep{meng2021cross} proposes a federated spatio-temporal model that leverages GNN-based architecture to encode the graph structure, constrained by cross-node federated learning, which disentangles temporal and spatial dynamics while reducing communication costs. While previous spatio-temporal analysis methods based on GNNs have made notable contributions, our \method{} stands out by addressing critical limitations and providing superior performance in capturing complex spatio-temporal relationships and diverse transition patterns in traffic data, these innovations lead to more accurate and robust traffic flow predictions.

\subsection{Traffic Flow Forecasting}
Traffic Flow Forecasting is a well-recognized and highly extensively researched problem in the field, garnering significant attention and interest~\citep{li2024survey}. Numerous spatio-temporal forecasting methods have been adapted for this task, yielding remarkable results. The predominant approach in addressing this challenge is rooted in machine learning algorithms, which leverage spatio-temporal data collected from an array of sensors to predict future traffic conditions. Traditional methods such as k-nearest neighbors~\citep{luo2019spatiotemporal}, autoregressive integrated moving average~\citep{williams2003modeling}, and support vector machines~\citep{sun2014traffic} have been employed, however, they often fall short in effectively modeling complex spatial relationships inherent in traffic data. With the rapid advancements in deep neural networks, deep learning-based methods have emerged as the dominant paradigm, focusing on the intricate modeling of spatio-temporal dependencies within traffic flow data~\citep{pan2019urban}. The fundamental idea revolves around utilizing deep neural architectures to capture these dependencies. This entails harnessing the power of GNNs to extract structured spatial relations encoded within road networks. Concurrently, sequence neural networks excel in capturing temporal dependencies over time. These two complementary approaches are often integrated to develop comprehensive models capable of handling the multifaceted intricacies of traffic flow forecasting~\citep{guo2019attention,zheng2020gman, wu2020connecting,li2023dynamic}. For instance, ASTGCN~\citep{guo2019attention} models recent, daily-periodic, and weekly-periodic traffic dependencies through spatial-temporal attention mechanisms, graph convolutions for spatial patterns, and standard convolutions for temporal features, with fused outputs for predictions. GMAN~\citep{zheng2020gman} introduces a graph multi-attention network that predicts future traffic conditions on a road network graph using an encoder-decoder architecture with spatio-temporal attention blocks. DGCRN~\citep{li2023dynamic} leverages hyper-networks to extract dynamic node attributes and generates dynamic filters at each time step for filtering node embeddings. However, existing spatio-temporal GNN methods still have some limitations in capturing high-order relationships and diverse transitional properties. To tackle this, we propose a novel method named \method{}, which not only conjointly explores high-order spatio-temporal correlations in constructed heterogeneous graphs extracted from both prior and posterior information, but also incorporates a conjoint self-attention decoder that leverages both multi-rank and multi-scale self-attention to capture diverse temporal transitional patterns.

\section{Methodology}
\begin{figure*}
    \centering \includegraphics[width=1\textwidth]{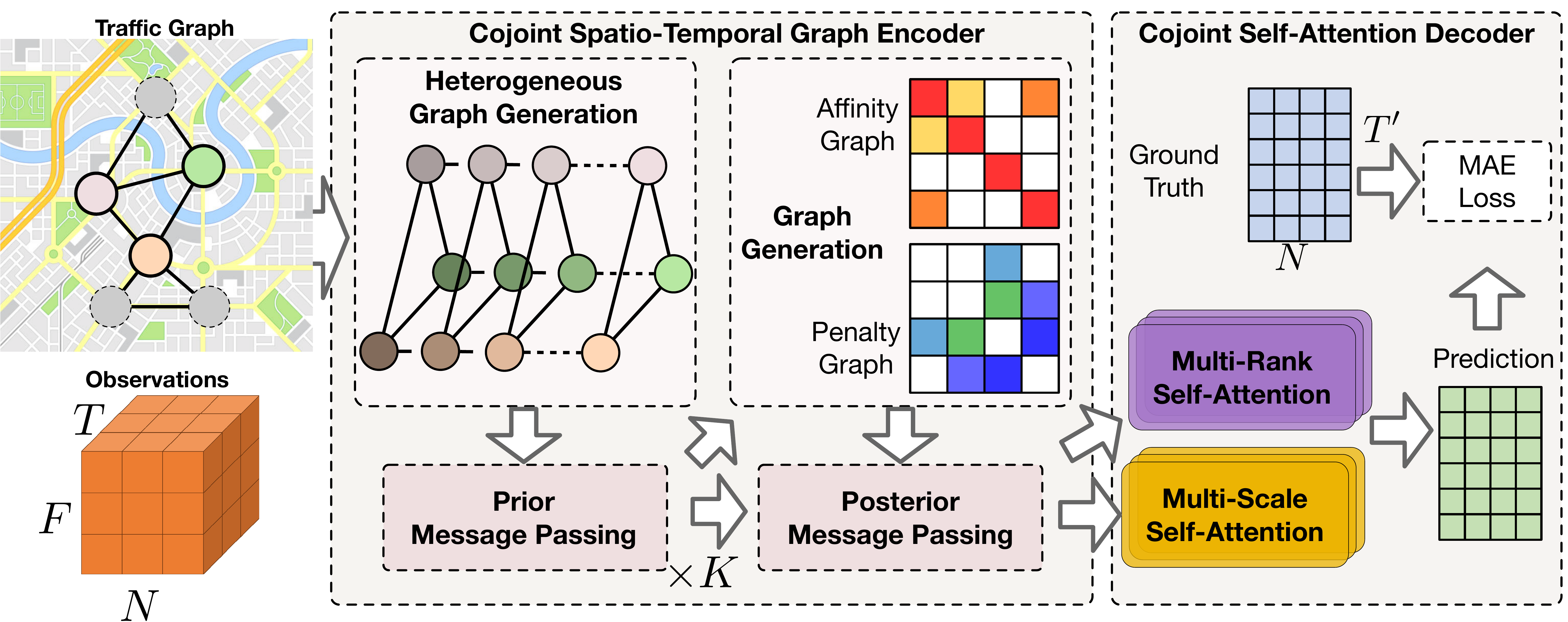}
    \caption{Framework overview of the proposed \method{}.} 
    \label{fig:framework}
\end{figure*}

This paper presents a new approach named \method{} for traffic flow forecasting. As illustrated in Figure \ref{fig:framework}, \method{} consists of a conjoint spatio-temporal graph encoder and a conjoint self-attention decoder. In the encoder, we conjointly extract high-order spatio-temporal correlations from both prior and posterior information. On the one hand, we perform message passing under the guidance of constructed heterogeneous graphs containing both prior spatial and temporal connections. On the other hand, we construct both semantic affinity graphs and penalty graphs to characterize dynamic relationships, and then incorporate both similarities and dissimilarities into sequential node representations. In the decoder, we aggregate sequential representations by exploring diverse transitional patterns from both multi-rank and multi-scale views. 

\smallskip\noindent\textbf{Problem Definition.}
We denote the traffic graph of road network as $\mathcal{G} = (\mathcal{V},\mathcal{E})$ with the node set $\mathcal{V}$ and the edge set $\mathcal{E}$. The adjacency matrix can be written as $A\in \mathbb{R}^{N \times N}$. The historical observation can be denoted as $\{\bm{X}_1,\bm{X}_2,\cdots,\bm{X}_T\}$ where $\bm{X}_t\in \mathbb{R}^{N\times F}$ is the observation at $t$ time step, $F$ is the dimension of each observation. The objective of traffic flow forecasting is to predict the future observations $\bm{X}_t(t>T)$. 

\subsection{Conjoint Spatio-temporal Graph Encoder}
The encoder is comprised of two components. It first constructs heterogeneous graphs for prior message passing and then infers affinity and penalty graphs for posterior message passing. 

\smallskip\noindent\textbf{Heterogeneous Graph Generator.} To learn composite spatio-temporal relationships, we present a heterogeneous graph that connects observations from both temporal and spatial views.

Specifically, we consider $r$ time steps in each heterogeneous graph $\mathcal{G}^H_{[t-r+1:t]}$, containing $rN$ observations $\{v_i^t\}_{t\in[t-r+1:t], v_i\in \mathcal{V}}$. The two observations are connected using spatial edges, i.e., $w_{v_i^t,v_j^t} = A_{ij}$ at each time step while consecutive observations are also connected using temporal edges, i.e., $w_{v_i^t,v_{i}^{t+1}} = 1$.

\smallskip\noindent\textbf{Prior Message Passing.} Then, we introduce prior message passing to capture composite correlations in traffic data. In particular, we leverage the message passing mechanism where each representation is updated by aggregating information from its neighborhood~\citep{gilmer2017neural}. Formally, the representation of $v_i^t$ at the $k$-th layer $\bm{h}_i^{t,(k)}$ can be written as:
\begin{equation}
\label{gnn}
\bm{h}_i^{t,(k)}= \operatorname{C}^{(k)}_{\theta}\left(\bm{h}_i^{t,(k-1)}, \operatorname{A}^{(k)}_{\theta} \left(\left\{\bm{h}_j^{t',(k-1)}\right\}_{v_j^{t'} \in \mathcal{N}(v_i^t)}\right) \right),
\end{equation}
where $\mathcal N(v_i^t)$ denote the neighbors of $v_i^t$. $\operatorname{A}^{(k)}_{\theta}$ and $\operatorname{C}^{(k)}_{\theta}$ represent the aggregation and combination operations parameterized by $\theta$ at the $k$-th layer, respectively. The final embedding of the node $v_i$ at time step $t$ and the $K$-th layer can be written as $\bm{h}_i^t = \bm{h}_i^{t,(K)} $.

\smallskip\noindent\textbf{Affinity and Penalty Graph Generator.} 
Nevertheless, prior message passing merely considers the affinities between pairs of nodes and often assigns zero weight to characterize dissimilarity~\citep{song2020spatial,li2023dynamic}. However, it is intuitively beneficial to account for the dissimilarity between nodes since it can signify complementary relationships among nodes, bearing significant implications for traffic condition modeling. To illustrate this, consider a traffic network where dissimilar nodes, such as a major highway and a narrow alley or a bustling urban intersection and a quiet suburban street, may exhibit contrasting traffic patterns. Capturing this dissimilarity can provide valuable insights into predicting traffic conditions accurately. Unfortunately, for each node, its dissimilar nodes are typically not participating in message passing, which could result in information loss and hence inferior performance. To tackle this issue, we provide both semantic affinity graphs and semantic penalty graphs to model diverse relationships in traffic networks.

In detail, we first calculate the correlation scores between node pairs using cosine similarity. Formally, given the embeddings after neighboring aggregation on the road network $\bm{h}^{t}_{i}$, we have:
\begin{equation}
\label{road_network}
s^t_{i j}=Score(\bm{h}^{t}_{i}, \bm{h}^{t}_{j})=\phi\left(\bm{w} \odot \bm{h}^{t}_{i}, \bm w \odot \bm{h}^{t}_{j}\right),
\end{equation}
where $\bm{w}$ is a learnable vector to decide the importance of different dimensions and $\odot$ is the Hadamard product. $\phi(\cdot, \cdot)$ calculates the cosine similarity between two vectors. Note that the correlation scores could be positive or negative. On this basis, we construct a semantic affinity graph $\bm{W}^t$ and a semantic penalty graph $\bm{P}^t_{i j}$ at the $t$-th step. Formally, 
\begin{equation}\label{Wij}
    \bm{W}^t_{i j}=\left\{\begin{array}{ll} s^t_{ij}  &  s_{i j} \geq 0 \\ 0, & \text { otherwise }\end{array}\right.
\end{equation}
\begin{equation}\label{Pij}
    \bm{P}^t_{i j}=\left\{\begin{array}{ll} -s^t_{ij}  &  s_{i j} < 0 \\ 0, & \text { otherwise }\end{array}\right.
\end{equation}

Similarly, we construct the heterogeneous version of two graphs containing observations between the time span $[t-r+1, t]$ by aggregating the temporal information. 
In formulation, we have the following equations:
\begin{equation}
\hat{\bm{W}}(v_{i}^{t},v_j^{t'})=\left\{\begin{array}{ll} Score(\bm{h}^{t}_{i}, \bm{h}^{t'}_{j}) & Score(\bm{h}^{t}_{i}, \bm{h}^{t'}_{j})>0 \\ 0 & \text{ otherwise }\end{array}\right.
\end{equation}
\begin{equation}
\hat{\bm{P}}(v_{i}^{t},v_j^{t'})=\left\{\begin{array}{ll} -Score(\bm{h}^{t}_{i}, \bm{h}^{t'}_{j}) & Score(\bm{h}^{t}_{i}, \bm{h}^{t'}_{j})<0 \\ 0 & \text{ otherwise }\end{array}\right.
\end{equation}

Note that the affinity graphs and penalty graphs also play a crucial role in handling noise in graph construction to ensure robustness. (i) Affinity Graphs $\bm{W}^t_{i j}$: These graphs capture positive correlations between node pairs based on cosine similarity scores. Nodes exhibiting similarity in traffic patterns are connected with non-zero weights, emphasizing their positive relationships. This helps in preserving meaningful connections in the graph. (ii) Penalty Graphs $\bm{P}^t_{i j}$: These graphs focus on negative correlations between node pairs, nodes with dissimilar traffic patterns are connected with non-zero weights representing the dissimilarity. This is essential for explicitly considering contrasting relationships and preventing the loss of information due to dissimilar nodes. By incorporating both types of graphs in the model, the system addresses the limitations of traditional message passing that often neglects dissimilarity. This explicit modeling of positive and negative relationships enhances the robustness of the graph representation to noise, ensuring that the model can effectively capture the diverse dynamics of a traffic network. It enables a more accurate and resilient traffic condition prediction by considering both complementary and contrasting relationships among nodes.

\smallskip\noindent\textbf{Posterior Message Passing.}
Intuitively, the node representations connected in the affinity graph should be close while the ones connected in the penalty graph should be far away. To accomplish this, we propose a correlation learning optimization objective as auxiliary loss function as follows:
\begin{equation}\label{IB1}
\begin{aligned}  
\mathcal{L}_{cor} = & \sum_{{t'}=t-r+1}^t\sum_{i=1}^N\sum_{v_j^{t''}\in N_W(v_i^{t'})} \hat{W}(v_{i}^{{t'}},v_j^{t''}) \gamma(\bm{u}_i^{t'}, \bm{h}_j^{t''})\\ &- \sum_{t'=t-r+1}^t\sum_{i=1}^N\sum_{v_j^{t''}\in N_P(v_i^{t'})} \hat{P}(v_{i}^{t'},v_j^{t''})\gamma(\bm{u}_i^{t'}, \bm{h}_j^{t''}) \\ &+ \beta \sum_{t'=t-r+1}^t\sum_{i=1}^N \gamma(\bm{u}^{t'}_i, \bm{h}^{t'}_i),
\end{aligned}
\end{equation}
where $\{\bm{u}^{t'}_i\}_{t'\in [t-r+1, t], i \in V}$ denotes the node representations to be optimized and $\gamma$ is a distance metric in the embedding space, $\beta$ is a hyperparameter used to balance the contributions of different losses (set to a default value of 1 in our experiments). The first term minimizes the distance between nodes connected in the affinity graph while the second term plays an opposite role for nodes connected in the penalty graph. The last term aims to reduce the variance of node representations for model stability. To facilitate optimization, $\gamma(\cdot, \cdot)$ is set to $l_2$-norm. Here, Eq.~\ref{IB1} has a \emph{closed solution} by calculating the partial derivatives and the optimal node representations with normalization ${\bm{u}^{t'}_i}^*$ can be derived as:
\begin{equation}
\label{posterior}  
{\bm{u}^{t'}_i}^* = \frac{\bm{h}^{t'}_i+ \sum_{v_j^{t''}} \hat{W}(v_{i}^{t'},v_j^{t''}) \bm{h}_j^{t''}- \sum_{v_j^{t''}} \hat{P}(v_{i}^{t'},v_j^{t''})\bm{h}_j^{t''}}{||\bm{h}^{t'}_i+ \sum_{v_j^{t''}} \hat{W}(v_{i}^{t'},v_j^{t''}) \bm{h}_j^{t''}- \sum_{v_j^{t''}} \hat{P}(v_{i}^{t'},v_j^{t''})\bm{h}_j^{t''}||_2}.
\end{equation}
Through posterior message passing, we have incorporated similarities and dissimilarities into node representations for effective traffic forecasting. Finally, for each node, the optimal node representations are concatenated into a tensor $\bm{U}_i=[\bm u_i^1,\cdots, \bm u_i^T]\in\mathbb{R}^{T\times d}$, where $d$ is the embedding dimension of the node representations.

\subsection{Conjoint Self-attention Decoder}
In reality, different locations in the traffic network could exhibit diverse transitional properties. For example, some locations could have periodic patterns like daily or monthly routines. To describe these diverse patterns in the traffic network, we offer a novel conjoint self-attention decoder, which sufficiently explores long-term correlations from both multi-rank and multi-scale perspectives.

\smallskip\noindent\textbf{Multi-rank Self-Attention Branch.} Recently, Transformer has been extensively used to explore large-scale data in deep learning~\citep{xia2022self,yang2022multi,zou2022improving,liu2021transformer,he2021transrefer3d,zhao2022target}. Inspired by this, we propose to utilize self-attention to identify long-term temporal relationships in traffic data. To decrease the parameters and avoid overfitting, we seek to utilize low-dimensional query vectors and value vectors for the low-rank attention matrix. Moreover, since various ranks may explore different features such as variances, our self-attention branch involves multiple ranks to generate multiple sequence embeddings. 

In detail, we take a given rank $\mu$ as an example. In this part, we omit the subscript of $\bm{U}_i$ since spatial correlations are not considered. In this branch, for each node, the representation $\bm u^t$ is converted into a query vector and a key vector, and their dot product is adopted to measure the importance of semantics at the current time step.
To compress the embedding matrix, we introduce two low-dimensional left transformation matrices $\tilde{\bm{K}}^{\mu}$ and $\tilde{\bm{V}}^{\mu} \in \mathbb{R}^{(T/r)\times T}$, where $r$ is the number of the heads. Following the paradigm in Transformer~\citep{vaswani2017attention}, three extra right transformation matrices $\mathcal{Q}^{\mu},\mathcal{K}^{\mu}$ and $\mathcal{V}^{\mu} \in \mathbb{R}^{d\times d}$ are defined to generate a query, a key and a value matrix. In formulation, we have:
\begin{equation}
\bm{\Gamma}^{\mu}=\operatorname{softmax}\left(\frac{\bm{U} \cdot \mathcal{Q}^{\mu} \left(\tilde{\bm{K}}^{\mu} \cdot \bm{U} \cdot \mathcal{K}^{\mu}\right)^{\top}}{\sqrt{d}}\right) \cdot \tilde{\bm{V}}^{\mu} \cdot \bm{U} \cdot \mathcal{V}^{\mu},
\end{equation}
where $\bm{\Gamma}^{\mu}=[\bm{\gamma}^{\mu,1},\cdots,\bm{\gamma}^{\mu,T}] \in \mathbb{R}^{T \times d}$. Finally, we summarize the embeddings for all time steps, producing rank-specific sequence embeddings $\bm{\gamma}^{\mu}\in \mathbb{R}^d$ and mean-pooling is adopted here. Similarly, for different ranks, we can obtain various rank-specific sequence embeddings. In our implementation, we choose three different ranks $\mu_1$, $\mu_2$ and $\mu_3$, which produce three sequence embeddings, i.e., $\bm{\gamma}^{\mu_1}$, $\bm{\gamma}^{\mu_2}$ and $\bm{\gamma}^{\mu_3}$ respectively.

\smallskip\noindent\textbf{Multi-scale Self-attention Branch.} Furthermore, taking into account the potential periodic patterns in different locations, we propose a multi-scale self-attention branch. This branch involves pooling representation sequences at different scales and then leverages the self-attention mechanism to effectively fuse these subsequence embeddings. 

Similarly, for each window size $\epsilon$ and each node, we can obtain subsequence embeddings by $\bm{\delta}^k=Pool(\bm{u}^{k\epsilon-\epsilon+1},\cdots,\bm{u}^{k\epsilon})$. Afterwards, the stacked matrix $\bm{\Delta}^\epsilon=[\bm{\delta}^1,\cdots, \bm{\delta}^{T/\epsilon}]\in \mathbb{R}^{T/\epsilon \times d}$ is fed into a self-attention module, which aggregates them into a scale-aware sequence embedding. Here, $\mathcal{Q}^{\epsilon},\mathcal{K}^{\epsilon}$ and $\mathcal{V}^{\epsilon} \in \mathbb{R}^{d\times d}$ 
denotes transformation matrices of a query, a key and a value, respectively.
Then, we calculate the weight using the dot-product to quantify the importance of each subsequence, and combine these subsequence representations to obtain a global representation. Formally, we have:
\begin{equation}
\bm{\Gamma}^{\epsilon}=\operatorname{softmax}\left(\frac{\bm{\Delta}^\epsilon \cdot \mathcal{Q}^{\epsilon} \left(\bm{\Delta}^\epsilon \cdot \mathcal{K}^{\epsilon}\right)^{\top}}{\sqrt{d}}\right) \cdot \bm{\Delta}^\epsilon \cdot \mathcal{V}^{\epsilon}.
\end{equation}
Similarly, with mean-pooling operation along the time dimension, we can generate the final global representation $\bm{\gamma}^{\epsilon}$.
Again, three window sizes, i.e., $\epsilon_1$, $\epsilon_2$ and $\epsilon_3$ are selected, generating three embeddings from different views $\bm{\gamma}^{\epsilon_1}$, $\bm{\gamma}^{\epsilon_2}$ and $\bm{\gamma}^{\epsilon_3}$, respectively.

\begin{algorithm}[!t]
\caption{The overall learning algorithm of \method{}.}
\label{alg1}
\begin{algorithmic}[1]
\REQUIRE Traﬃc graph of road network $\mathcal{G}$, traffic signals over the past time steps $\bm{X}$;
\ENSURE Prediction of traffic signals in future time steps;
\STATE Construct heterogeneous graph via Heterogeneous Graph Generator in Sec. 3.1;
\REPEAT
\STATE Compute each prior node representation $\bm{h}_i^t$ by Eq.~\ref{gnn};
\STATE Construct affinity and penalty graphs by Eq.~\ref{Wij} and ~\ref{Pij};\STATE Compute each posterior node representation $\bm{u}_i^t$ by Eq.~\ref{posterior};
\STATE Compute global embeddings $\bm{g}$ by Eq.~\ref{fusion};
\STATE Output the final prediction and calculate the MAE loss by Eq.~\ref{mse};
\STATE Backpropagation and update parameters by gradient descend;
\UNTIL convergence
\end{algorithmic}
\end{algorithm}

Finally, we introduce learnable parameters to aggregate these learned embeddings. In particular, with $\{w^{\mu_{j}}\}_{j=1}^3$ and $\{w^{\epsilon_{j}}\}_{j=1}^3$, and we can generate the final embedding $\bm{g}$:
\begin{equation}
    \label{fusion}
    \bm{g} =\frac{\sum_{j=1}^3\exp(w^{\mu_{j}})\bm{\gamma}^{\mu_j}+\exp(w^{\epsilon_j})\bm{\gamma}^{\epsilon_j}}{\sum_{j=1}^3\exp(w^{\mu_{j}})+\exp(w^{\epsilon_j}) }.
\end{equation}
The embedding $\bm{g}$ would be combined with the state of the final step, i.e., $\bm{u}^T$ to generate the predictions using a multi-layer perceptron (MLP) as:
\begin{equation}
\hat{Y}=\text{MLP}\left(\bm{g}||\bm{u}^T\right),
\end{equation}
where $\hat{Y}$ is the prediction, $||$ denotes the concatenation operation.
We optimize the overall framework by minimizing the standard mean absolute error (MAE) loss as follows:
\begin{equation}
\label{mse}
\mathcal{L}_{mse}=\sum_{t}\left|Y_t-\hat{Y}_t\right|.
\end{equation}
The whole algorithm is summarized in Algorithm \ref{alg1}.

\section{Experiments}
\subsection{Experimental Setup}


\begin{table}
    \centering
    \caption{Statistics of the evaluation datasets.}
    \label{tab:dataset}
    \resizebox{0.49\textwidth}{!}{
    \begin{tabular}{cccccc}
    \toprule
        \textbf{Dataset}     & PEMS-BAY  & PEMS08 & METR-LA & PEMS07 \\ \midrule
        $|\mathcal{V}|$ & 325 & 170 & 207 & 883 \\
        $|\mathcal{E}|$ & 2369 & 295 & 1515 & 866 \\
        \textbf{Time Steps} & 52,116 & 17,856 & 34,272 & 28,224 \\
        \multirow{2}{*}{\textbf{Time Range}} & 01/01/2017-  & 07/01/2016-  & 03/01/2012-  & 05/01/2017- \\ & 03/31/2017 & 08/31/2016 & 06/30/2012 & 08/31/2017\\
        \textbf{Interval} & 5min & 5min & 5min & 5min \\
        \multirow{2}{*}{\textbf{Place}} & San Francisco  & San Bernardino  & Los Angeles,  & California, \\ & Bay Area, USA & Area, USA & USA & USA\\
    \bottomrule
    \end{tabular}
    }
\end{table}

\smallskip
\noindent\textbf{Datasets and Metrics.} To comprehensively assess the performance of our proposed \method{}, we conduct experiments on four real-world traffic datasets: \textit{PEMS-BAY}, \textit{PEMS08}, \textit{METR-LA}, and \textit{PEMS07}, and we summarize the statistics of datasets in Table~\ref{tab:dataset}.

For each of these datasets, we utilize sixty minutes of historical data to forecast traffic conditions for the subsequent sixty minutes. To evaluate the forecasting accuracy, we employ three common metrics: mean absolute error (MAE), mean absolute percentage error (MAPE), and root mean squared error (RMSE), considering prediction horizons of 3, 6, and 12 time steps, providing a comprehensive assessment of \method{}'s predictive capabilities across different datasets and forecasting periods.

\smallskip\noindent \textbf{Baseline Models.} To comprehensively evaluate the performance of our proposed \method{}, we benchmark it against a diverse set of baseline models, encompassing both traditional methods and state-of-the-art neural network-based methods. Traditional methods involve:

\begin{itemize}
\item HA (Historical Average): HA is a baseline method for prediction that utilizes past data averages as a simple forecasting strategy.

\item VAR (Vector Auto-Regressive)~\citep{lutkepohl2005new}: VAR is a time series model that extends autoregression to multiple variables, capturing interdependencies among them.

\item SVR (Support Vector Regression)~\citep{wu2004travel}: SVR employs a linear support vector machine for regression in the domain of classical time series analysis.
\end{itemize}

Neural network-based methods comprise:
\begin{itemize}
\item DCRNN~\citep{li2017diffusion}: DCRNN models traffic flow as a diffusion process on a directed graph, utilizing bidirectional random walks on the graph and an encoder-decoder architecture with scheduled sampling to capture spatiotemporal dependencies.

\item STGCN~\citep{yu2017spatio}: STGCN employs a fully convolutional structure with a combination of graph convolutional layers and convolutional sequence learning layers to model spatial and temporal dependencies. 

\item ASTGCN~\citep{guo2019attention}: ASTGCN adopts the spatio-temporal attention mechanism to capture the spatio-temporal correlation in traffic flow, and leverages graph convolution and standard convolution to mine static spatio-temporal features. 

\item STSGCN~\citep{song2020spatial}:  STSGCN designs spatial-temporal synchronous modeling mechanism and multiple modules for different time periods to capture the localized correlations and heterogeneities. 

\item MTGNN~\citep{wu2020connecting}: MTGNN proposes a novel mix-hop propagation layer and a dilated inception layer to capture the spatial and temporal dependencies, with the function of automatically extracting the uni-directed relations.

\item GMAN~\citep{zheng2020gman}: GMAN incorporates multiple spatio-temporal attention blocks into the encoder-decoder architecture to model the impact of the spatio-temporal factors. 

\item DGCRN~\citep{li2023dynamic}: DGCRN leverages filtered node embeddings to generate a dynamic graph and combines it with the pre-defined static graph, which jointly facilitates topology modeling. 

\item DSTGCN~\citep{liu2023spatial}: DSTGCN performs similarity learning to extract the complex spatial–temporal relationships and utilizes a convolution module with an attention mechanism to dynamically extract spatial–temporal dependence. 

\item PDFormer~\citep{jiang2023pdformer}: PDFormer proposes a spatial self-attention module and a traffic delay-aware feature transformation module to respectively model dynamic spatial dependencies and the time delay of spatial information propagation. 

\item STAEformer~\citep{liu2023spatio}: STAEformer designs the spatio-temporal adaptive embedding to capture the intricate spatiotemporal traffic patterns and address the diminishing performance gains.
\end{itemize}



\begin{table*}[!t]
\renewcommand\arraystretch{0.91}
    \centering
    \setlength{\abovecaptionskip}{2mm}
    \setlength{\belowcaptionskip}{1mm}
    \caption{
        Traffic forecasting results on the PEMS-BAY, PEMS08 and METR-LA datasets.}
    \label{tab:main}
    \resizebox{1\textwidth}{!}{
    \begin{tabular}{ccccr|ccr|ccr}
        \toprule
        \multirow{2}*{\textbf{Datasets}} &\multirow{2}*{\textbf{Methods}} & \multicolumn{3}{c}{\textbf{Horizon 3}} & \multicolumn{3}{c}{\textbf{Horizon 6}}& \multicolumn{3}{c}{\textbf{Horizon 12}}\\ 
        \cmidrule(r){3-5} \cmidrule(r){6-8} \cmidrule(r){9-11}
        &  & MAE & RMSE & MAPE & MAE & RMSE & MAPE & MAE & RMSE & MAPE\\
        \midrule
    \multirow{14}*{\textbf{PEMS-BAY}} 
        &HA              & 1.89  & 4.30  & 4.16\%        & 2.50  & 5.82  & 5.62\%       & 3.31  & 7.54  & 7.65\% \\ 
        &VAR             & 1.74  & 3.16  & 3.60\%        & 2.32  & 4.25  & 5.00\%       & 2.93  & 5.44  & 6.50\% \\ 
        &SVR             & 1.85  & 3.59  & 3.80\%        & 2.48  & 5.18  & 5.50\%       & 3.28  & 7.08  & 8.00\% \\ 
        &DCRNN           & 1.38  & 2.95  & 2.90\%        & 1.74  & 3.97  & 3.90\%       & 2.07  & 4.74  & 4.90\% \\ 
        &STGCN           & 1.36  & 2.96  & 2.90\%        & 1.81  & 4.27  & 4.17\%       & 2.49  & 5.69  & 5.79\% \\ 
        &ASTGCN          & 1.52  & 3.13  & 3.22\%        & 2.01  & 4.27  & 4.48\%       & 2.61  & 5.42  & 6.00\% \\  
        &STSGCN          & 1.44  & 3.01  & 3.04\%        & 1.83  & 4.18  & 4.17\%       & 2.26  & 5.21  & 5.40\% \\  
        &MTGNN           & 1.32  & 2.79  & 2.77\%        & 1.65  & 3.74  & 3.69\%       & 1.94  & 4.49  & 4.53\% \\  
        &GMAN            & 1.34  & 2.91  & 2.86\%        & 1.63  & 3.76  & 3.68\%       & 1.86  & 4.32  & 4.37\% \\  
        &DGCRN           & 1.28  & 2.69  & 2.66\%        & 1.59  & 3.63  & 3.55\%       & 1.89  & 4.42  & 4.43\% \\ 
        &DSTGCN         & 1.17  & 2.30  & 2.37\%        & 1.60  & 3.54  & 3.63\%       & 1.94  & 4.39  & 4.65\% \\ 
        &PDFormer         & 1.16  & 2.31  & 2.36\%        & 1.61  & 3.61  & 3.58\%       & 1.96  & 4.45  & 4.55\% \\ 
        &STAEformer         & 1.30  & 2.78  & 2.76\%       & 1.61  & 3.69  & 3.62\%       & 1.88  & 4.34  & 4.37\% \\ 
    \cmidrule(r){2-11}
    &\method{}      & \textbf{1.13}  & \textbf{2.23}  & \textbf{2.29\%}        & \textbf{1.53}  & \textbf{3.42}  & \textbf{3.38\%}      & \textbf{1.84}  & \textbf{4.23}  & \textbf{4.28\%} \\ 
    \midrule  
    \color{black}{\multirow{14}*{\textbf{PEMS08}}}
        &HA              & 23.52  & 34.96  & 14.72\%        & 27.67  & 40.89  & 17.37\%       & 39.28  & 56.74  & 25.17\% \\ 
        &VAR             & 19.52  & 29.73  & 12.54\%        & 22.25  & 33.30  & 14.23\%        & 26.17  & 38.97  & 17.32\% \\ 
        &SVR             & 17.93  & 27.69  & 10.95\%        & 22.41  & 34.53  & 13.97\%       & 32.11  & 47.03  & 20.99\% \\ 
        &DCRNN           & 15.64  & 25.48  & 10.04\%        & 17.88  & 27.63  & 11.38\%       & 22.51  & 34.21  & 14.17\% \\ 
        &STGCN           & 15.30  & 25.03  &  9.88\%        & 17.69  & 27.27  & 11.03\%       & 25.46  & 33.71  & 13.34\% \\ 
        &ASTGCN          & 16.48  & 25.09  & 11.03\%        & 18.66  & 28.17  & 12.23\%       & 22.83  & 33.68  & 15.24\% \\  
        &STSGCN          & 15.45  & 24.39  & 10.22\%        & 16.93  & 26.53  & 10.84\%       & 19.50  & 30.43  & 12.27\% \\  
        &MTGNN           & 14.24  & 22.43  &  9.02\%        & 15.30  & 24.32  &  9.58\%       & 16.85  & 26.93  & 10.57\% \\  
        &GMAN            & 13.80  & 22.88  &  9.41\%        & 14.62  & 24.02  &  9.57\%       & 15.72  & 25.96  & 10.56\% \\  
        &DGCRN           & 13.89  & 22.07  &  9.19\%        & 14.92  & 23.99  & 9.85\%       & 16.73  & 26.88  & 10.84\% \\  
        &DSTGCN         & 13.65  & 21.86  & 10.26\%        & 14.50  & 23.67  & 11.48\%       & 15.78  & 26.09  & 12.50\% \\ 
        &PDFormer          & 13.02  & 22.25  & 8.70\%        & 13.81  & 24.02  & 9.14\%       & 15.59  & 26.09  & 10.41\% \\
        &STAEformer          & \textbf{12.91}  & 22.17  & \textbf{8.61\%}        & \textbf{13.67}  & 23.71  & \textbf{9.00\%}       & 15.17 & 25.87  &10.08\% \\ 
    \cmidrule(r){2-11}
    &\method{}     & 13.18  & \textbf{21.87}  & 8.82\%        & 13.98  & \textbf{23.65}  & 9.18\%      & \textbf{15.07}  & \textbf{25.93}  & \textbf{9.81\%} \\ 
        \midrule 
            \multirow{14}*{\textbf{METR-LA}} 
        &HA              & 4.79  & 10.00 & 11.70\%       & 5.47  & 11.45 & 13.50\%      & 6.99  & 13.89  & 17.54\% \\ 
        &VAR             & 4.42  & 7.80  & 13.00\%       & 5.41  & 9.13  & 12.70\%      & 6.52  & 10.11 & 15.80\% \\ 
        &SVR             & 3.39  & 8.45  & 9.30\%        & 5.05  & 10.87 & 12.10\%      & 6.72  & 13.76 & 16.70\% \\ 
        &DCRNN           & 2.77  & 5.38  & 7.30\%        & 3.15  & 6.45  & 8.80\%       & 3.60  & 7.60  & 10.50\% \\ 
        &STGCN           & 2.88  & 5.74  & 7.62\%        & 3.47  & 7.24  & 9.57\%       & 4.59  & 9.40  & 12.70\% \\ 
        &ASTGCN          & 4.86  & 9.27  & 9.21\%        & 5.43  & 10.61 & 10.13\%      & 6.51  & 12.52 & 11.64\% \\  
        &STSGCN          & 3.31  & 7.62  & 8.06\%        & 4.13  & 9.77  & 10.29\%      & 5.06  & 11.66 & 12.91\% \\  
        &MTGNN           & 2.69  & 5.18  & 6.88\%        & 3.05  & 6.17  & 8.19\%       & 3.49  & 7.23  & 9.87\% \\  
        &GMAN            & 2.80  & 5.55  & 7.41\%        & 3.12  & 6.49  & 8.73\%       & 3.44  & 7.35  & 10.07\% \\  
        &DGCRN           & 2.62  & 5.01  & 6.63\%        & \textbf{2.99}  & 6.05  & \textbf{8.02}\%       & 3.44  & 7.19  & 9.73\% \\ 
        &DSTGCN         & 2.68  & 4.97  & 7.21\%        & 3.12  & 6.18  & 9.02\%       & 3.50  & 7.19  & 10.65\% \\ 
        &PDFormer         & 2.83  & 5.59  & 7.26\%        & 4.05  & 6.59  & 10.32\%       & 4.80  & 7.82  & 11.20\% \\ 
        &STAEformer         & 2.65  & 5.11  & 6.86\%        & 3.04  & 6.01  & 8.15\%      & 3.50  & 7.12  & 9.83\% \\ 
    \cmidrule(r){2-11}
    &\method{}      & \textbf{2.57}  & \textbf{4.70}  & \textbf{6.43\%}        & 3.01  & \textbf{5.96}  & 8.13\%      & \textbf{3.43}  & \textbf{7.07}  & \textbf{9.70\%} \\ 
        \bottomrule
    \end{tabular}
    }
    \end{table*}

\smallskip\noindent \textbf{Implemental Details.} 
Our proposed model \method{} is optimized on an NVIDIA RTX GPU, capitalizing on its parallel processing capabilities to accelerate training and inference tasks. For our proposed model, we set the embedding dimension $d$ to 64 for the encoder. The encoder comprises six prior message-passing layers, enabling the model to process and propagate information across the graph structure efficiently. For the decoder, we carefully tune two critical components: ranks $[\mu_1, \mu_2, \mu_3]$ and window sizes $[\epsilon_1, \epsilon_2, \epsilon_3]$. Specifically, we choose to use values of $[3, 4, 6]$ for both ranks and window sizes. This configuration was found to be optimal and demonstrated robust performance across different datasets and prediction horizons. At the end of the decoder, a two-layer fully connected neural network is employed to transform the learned representations into the final prediction. To optimize the entire framework, we utilize the Adam optimizer~\citep{kingma2014adam}, a widely used optimization algorithm that adapts learning rates during training. The model is trained for 100 epochs to ensure convergence, with a learning rate of 0.001 and a batch size of 32, all carefully chosen to balance training efficiency and model performance.

\begin{table*}[t]
\centering
\caption{Comparison of methods across different horizons (from H 1 to H 12). The experiments are performed on PEMS07 dataset and MAPEs are shown.}
\label{tab:horizon}
\resizebox{\textwidth}{!}{%
\begin{tabular}{lcccccccccccc}
\toprule
Method     & H 1 & H 2 & H 3 & H 4 & H 5 & H 6 & H 7 & H 8 & H 9 & H 10 & H 11 & H 12
    \\
\midrule
COOL       & 1.16\% & 7.27\% & 7.78\% & 7.72\% & 7.85\% & 8.03\% & 8.14\% & 8.31\% & 8.47\% & 8.60\% & 8.66\% & 8.76\%
        \\
STAEformer & 6.96\%       & 7.30\%       & 7.55\%       & 7.75\%       & 7.91\%       & 8.06\%       & 8.20\%       & 8.36\%       & 8.48\%       & 8.63\%       & 8.76\%       & 8.97\%       \\
PDFormer   & 7.99\%       & 8.25\%       & 8.53\%       & 8.70\%       & 8.94\%       & 9.09\%       & 9.14\%       & 9.37\%       & 9.46\%       & 9.66\%       & 9.84\%       & 10.42\%      \\
\bottomrule
\end{tabular}%
}
\end{table*}

\subsection{The Performance of \method{}}
In our experiments, we follow the standard dataset division protocols commonly followed in previous works~\citep{li2023dynamic}. To ensure robust evaluation, we employ the following data split strategies for each of the datasets under consideration: For the PEMS-BAY and METR-LA datasets, we allocated 70\% of the data for training, 10\% for validation, and the remaining 20\% for testing. For the PEMS08 dataset, we opt for a slightly different split, with 60\% of the data used for training, 20\% for validation, and the final 20\% for testing. The results of our experiments compared with a range of different baselines, are presented in Table~\ref{tab:main}. We also provide more detailed results on the PEMS07 dataset in Table \ref{tab:horizon}, where we present the results of 12 horizons. From the results, we have three observations as follows:
\begin{itemize}
    \item Generally, traditional methods showcase inferior performance when compared to cutting-edge neural network-based approaches. This performance gap can be attributed to the fact that traditional methods predominantly consider temporal correlations while neglecting critical spatial dependencies within the data.
    \item Our proposed \method{} achieves the best performance across a majority of settings and datasets, underscoring the effectiveness of our novel framework. Note that PDFormer~\citep{jiang2023pdformer} and STAEformer~\citep{liu2023spatio}, both based on the transformer mechanism, exhibit slightly superior performance on the PEMS08 dataset for Horizon=3 and 6 predictions. This may be attributed to the ability of spatiotemporal transformers to effectively capture dependencies over medium to long time intervals. However, as the time horizon increases (Horizon=12), the superiority of our approach in long-range predictions becomes more pronounced. This notable success can be primarily attributed to two key factors. First, our proposed \method{} leverages a conjoint spatio-temporal graph encoder, which facilitates the exploration of higher-order relations, making it adept at capturing complex dependencies within the traffic flow data. Second, our model incorporates multi-rank and multi-scale self-attention branches, enabling it to effectively capture a diverse range of sequential trends.
    \item To demonstrate the superiority of our proposed COOL across diverse horizons, we compare it with state-of-the-art baselines (i.e., PDFormer and STAEformer) on the PEMS07 dataset. We evaluate the performance using MAPEs as metrics for different horizons ranging from Horizon 1 to 12. The results, shown in Table \ref{tab:horizon}, reveal that our model consistently outperforms the competitive methods at all different horizons compared to the baselines. This clearly highlights the outstanding capability of our proposed model in exploring spatio-temporal dependencies.
    \item Remarkably, our model exhibit the most substantial improvement on the PEMS-BAY and PEMS07 datasets, which are among the largest and most challenging datasets considered. This underscores the scalability and robustness of our model when faced with large-scale traffic prediction tasks.
\end{itemize}

\begin{table}[!t]
\renewcommand\arraystretch{0.935}
    \centering
    \setlength{\abovecaptionskip}{2mm}
    \setlength{\belowcaptionskip}{1mm}
    \caption{
        Ablated study on PEMS08 and METR-LA datasets. The performance is measured in terms of MAE.}
    \label{tab:ablation}
    \resizebox{0.49\textwidth}{!}{
    \begin{tabular}{lccc}
        \toprule
        {{PEMS08}}  & {{Horizon 3}} & {{Horizon 6}}& {{Horizon 12}}\\
        \midrule
\method{} & \textbf{13.18} & \textbf{13.98} & \textbf{15.07} \\
w/o Prior & 13.21 & 14.04 & 15.26 \\
w/o Posterior & 13.26 & 14.09 & 15.19 \\
w/o Multi-rank & 13.37 & 14.28 & 15.43 \\
w/o Multi-scale & 13.34 & 14.10 & 15.25
    \\
    \midrule
    {{METR-LA}}  & {{Horizon 3}} & {{Horizon 6}}& {{Horizon 12}}\\
    \midrule
\method{} & \textbf{2.57} & \textbf{3.01} & \textbf{3.43} \\
w/o Prior & 2.64 & 3.10 & 3.51 \\
w/o Posterior & 2.60 & 3.05 & 3.46 \\
w/o Multi-rank & 2.59 & 3.04 & 3.45 \\
w/o Multi-scale & 2.59 & 3.03 & 3.44 \\
    \bottomrule
    \end{tabular}
    }
    \end{table}

\subsection{Ablation Studies}
In this subsection, we thoroughly conduct ablation studies to assess the efficacy of each constituent component incorporated within our proposed \method{} framework. Our evaluation is performed on both the PEMS08 and METR-LA datasets. To conduct these ablations, we systematically remove individual components, including prior graph convolution, posterior graph convolution, multi-rank self-attention, and multi-scale self-attention, from our \method{} model. Subsequently, we evaluate the performance of the modified models with these components removed.

The results of these ablation experiments are carefully reported in Table~\ref{tab:ablation}. As can be seen from the results, the removal of any single component inevitably leads to a noticeable performance degradation. This compelling evidence highlights the pivotal effectiveness of each module in our framework. It's noteworthy that removing either the prior graph convolution or the posterior graph convolution in isolation does not significantly impact the results. This suggests a degree of complementarity between these two graph convolution components, further illustrating the inherent robustness of the \method{} model.

\begin{figure}[t]
\centering
\setlength{\abovecaptionskip}{0.1cm}
\subfigure[]{
\label{fig:hyper:a}
\includegraphics[width=0.231\textwidth]{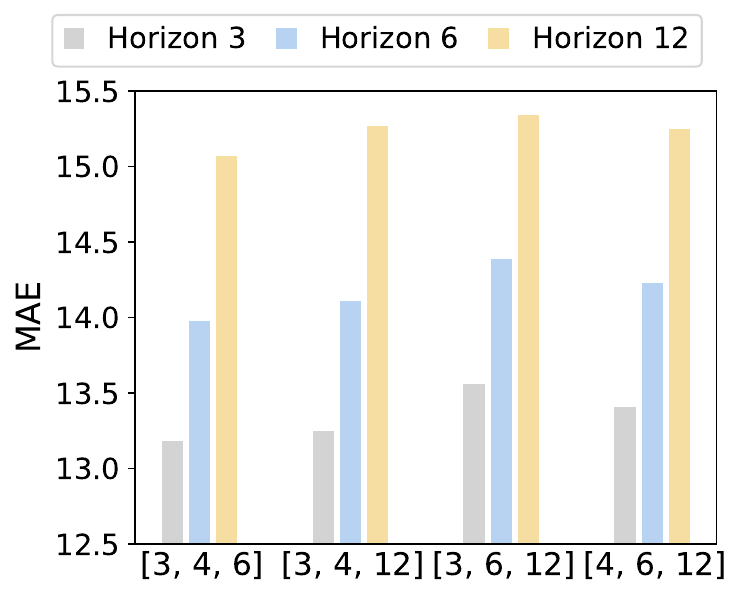}}
\subfigure[]{
\label{fig:hyper:b}
\includegraphics[width=0.231\textwidth]{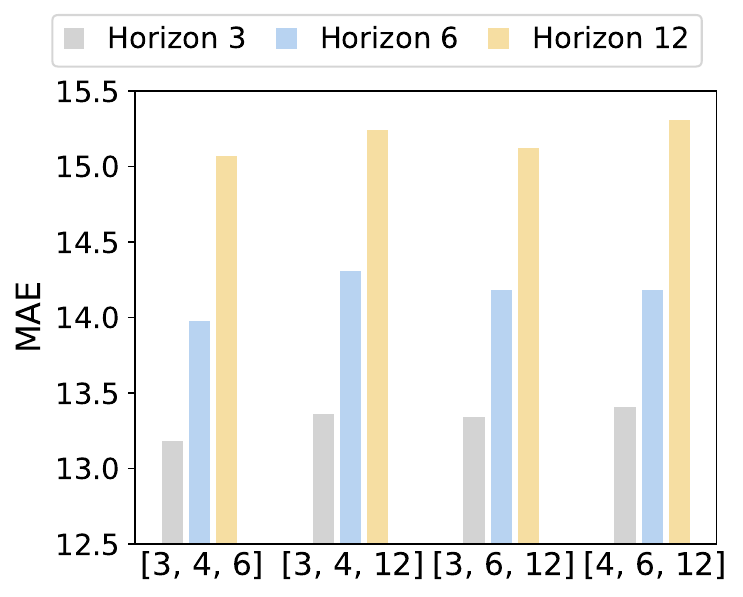}}
\subfigure[]{
\label{fig:hyper:c}
\includegraphics[width=0.231\textwidth]{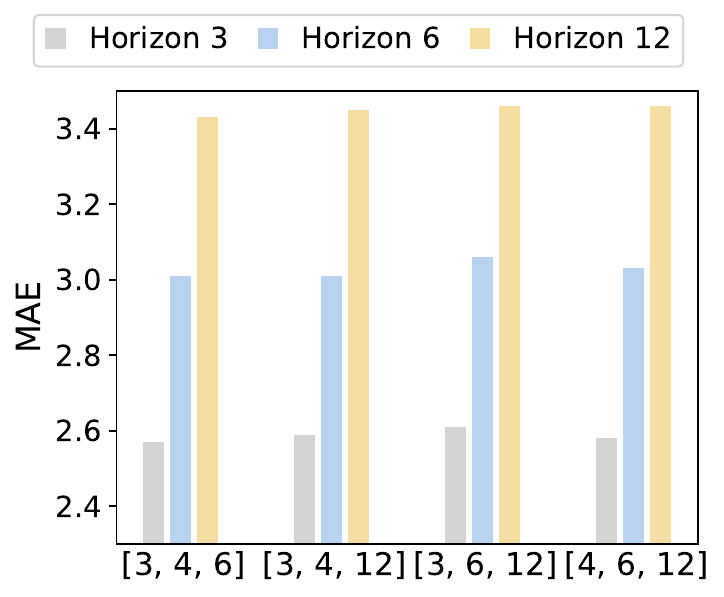}}
\subfigure[]{
\label{fig:hyper:d}
\includegraphics[width=0.231\textwidth]{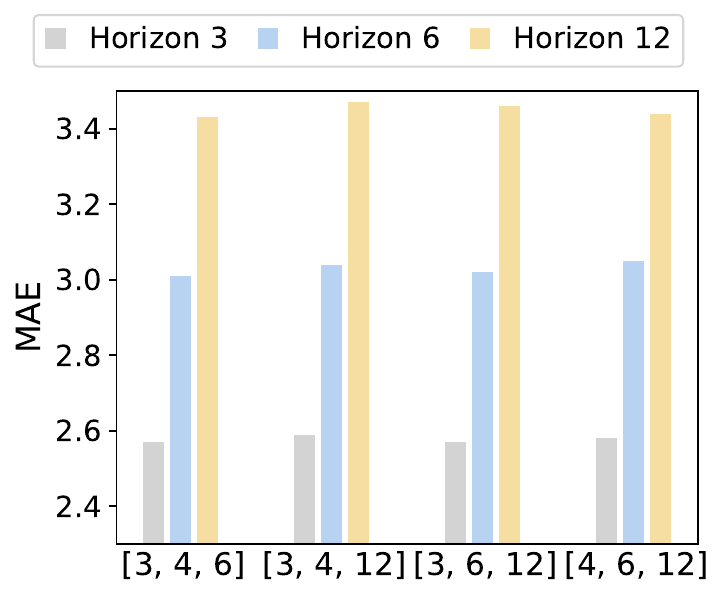}}
\caption{Hyperparameter study of the proposed \method{} on PEMS08 and METR-LA.
}
\label{fig:hyper}
\vspace{-0.5cm}
\end{figure}
\subsection{Hyperparameter Analysis}
In this part, we investigate the sensitivity of the model's hyperparameters, with a specific focus on the different ranks and window sizes employed in multi-rank and multi-scale self-attention mechanisms, respectively. Our findings are thoroughly presented in Figure~\ref{fig:hyper}, where we distinguish between experiments conducted on the PEMS08 dataset (depicted in Figure (a) and (b)) and those carried out on METR-LA (illustrated in Figure (c) and (d)). Figures (a) and (c) delve into the influence of hyperparameters $\mu_1$, $\mu_2$, and $\mu_3$ in the context of multi-rank self-attention, while Figures (b) and (d) analyze the impact of $\epsilon_1$, $\epsilon_2$, and $\epsilon_3$ within the realm of multi-scale self-attention. Generally, our proposed \method{} exhibits a robustness to variations in these hyperparameters. However, it is noteworthy that a combination of window sizes $[3, 4, 6]$ consistently yields slightly superior performance in both the multi-rank and multi-scale self-attention branches. This phenomenon might be attributed to the fact that medium-sized window sizes exhibit an enhanced capability to capture the subtle variations in traffic conditions, as opposed to larger window sizes.

\subsection{Efficiency Analysis}
To evaluate our model's performance efficiency against state-of-the-art baseline models. Specifically, we compare our model against PDFormer and STAEformer in terms of the number of parameters, the duration of training per epoch, and the testing time. The results clearly demonstrate that our model COOL operates with a significantly lower number of parameters, indicating a higher efficiency in terms of model size compared to the baselines. Besides, COOL's training time is comparable to that of the baseline models, showcasing its efficiency in learning. Moreover, our COOL achieves a shorter test time than both baselines, underscoring its superior performance in terms of inference speed and operational efficiency. This highlights COOL's effectiveness in balancing model complexity with computational efficiency, making it a highly competitive choice for applications requiring rapid predictions.

\begin{table}[!t]
\centering
\caption{Comparison of model efficiency}
\label{tab:model_efficiency}
\resizebox{0.48\textwidth}{!}{%
\begin{tabular}{lccc}
\toprule
Model & Number of Parameters & Training Time per Epoch & Test Time \\
\midrule
COOL& 264K & 87.3s & 5.4s \\
PDFormer & 531K & 86.9s & 9.3s \\
STAEFormer & 1123K & 91.0s & 6.6s \\
\bottomrule
\end{tabular}%
}
\end{table}

\begin{figure}[t]
\centering
\setlength{\abovecaptionskip}{0.1cm}
\subfigure[Visualization on sensor No.19]{
\label{fig:visualization:a}
\includegraphics[width=0.47\textwidth]{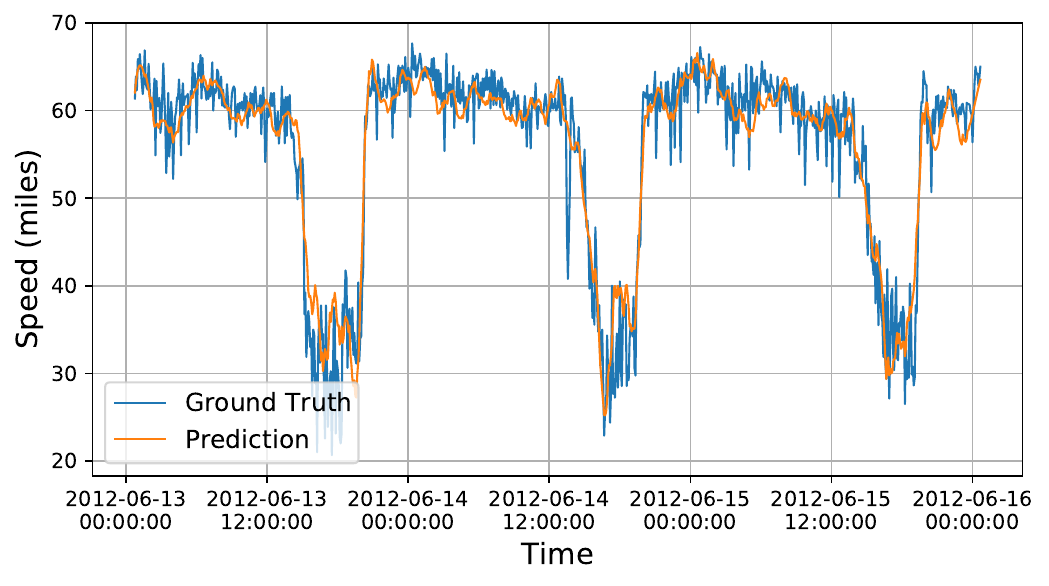}}
\subfigure[Visualization on sensor No.194]{
\label{fig:visualization:b}
\includegraphics[width=0.47\textwidth]{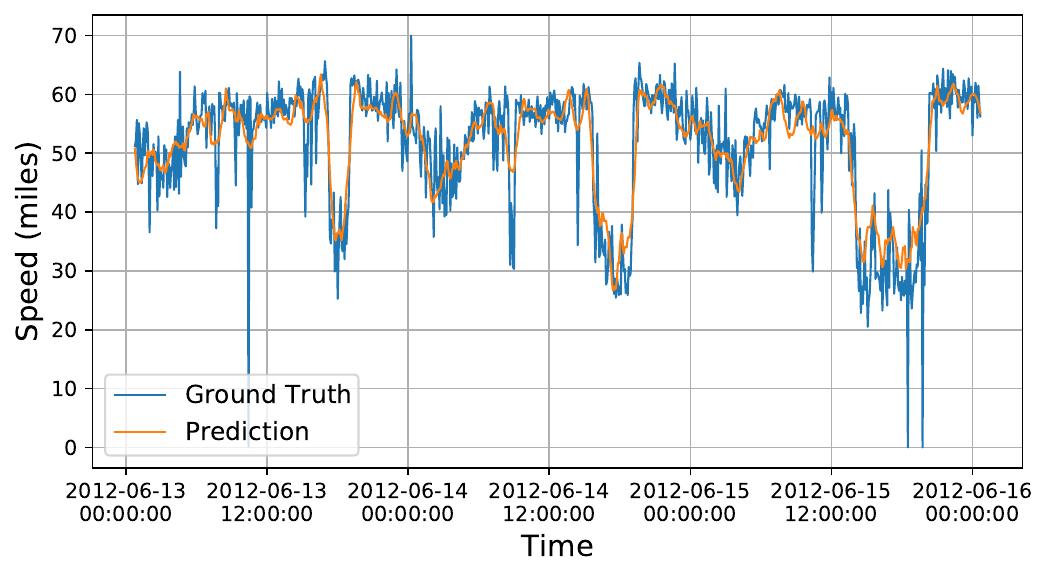}}
\caption{Visualization of prediction results on METR-LA.}
\label{fig:visualization}
\end{figure}

\subsection{Visualization of Prediction Results}
In this part, we provide the visualization of forecasting results. 
Figure~\ref{fig:visualization} visualizes the prediction results and ground truth values of sensors No.19 and No.194 from Jun 13 to Jun 15. Except for some noises (maybe caused by sudden accidents or sensor failures), the proposed \method{} predicts the traffic condition with impressive accuracy. In sensor No.19, the pattern is regular: every evening the traffic speed drops due to congestion. The model easily captures this pattern and provides a good prediction. In sensor No.194, the traffic condition is more diverse and complicated: the traffic speed drops significantly in the afternoon, but the starting time and duration of this congestion are very different over the three days. On the first day, this afternoon congestion appears late and is quickly over, whereas on the last day it comes early and lasts longer. By handling the diverse traffic conditions, our proposed \method{} successfully predicts the traffic speed with reasonable accuracy under this complicated situation, validating the superiority of our method. 

\begin{figure}
    \centering
    \includegraphics[width=\linewidth]{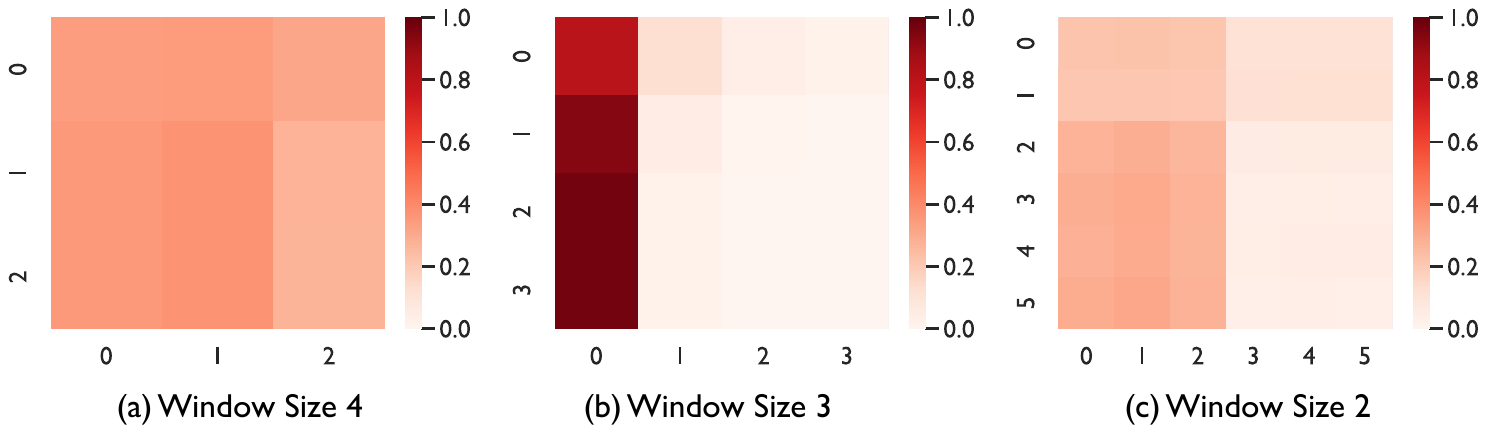}
    \caption{Visualization of learned attentions scores of multi-scale attention module. Sub-figure (a) is the attention scores of window size 4, Sub-figure (b) is the attention scores of window size 3, Sub-figure (c) is the attention scores of window size 2.}
    \label{fig:attn}
\end{figure}

\subsection{Visualization of Learned Attentions}
In this subsection, we provide the visualization of learned attention. More specifically, we visualize the attention scores multi-scale self-attention of the proposed \method{}. The experiments are performed on METR-LA dataset and the learned attentions of multiple scales are visualized using the heatmaps. The experimental results are shown in Figure \ref{fig:attn}. As can be seen from the results, the attention of different scales exhibits different patterns which shows that the proposed attention module can capture meaningful information at different scales. For example, in Figure \ref{fig:attn} (a), we use a window size of 4 and the attention distributed uniformly within the inputs. In comparison, when the window size is set to 3 (shown in Figure \ref{fig:attn} (b)), the first window captures meaningful information that is attended by other inputs. This shows that arranging the temporal inputs at different scales and learn the attended output thereafter is a way to view the data from different perspectives and are more likely to capture essential information at specific scales.

\begin{figure}
    \centering
    \includegraphics[width=\linewidth]{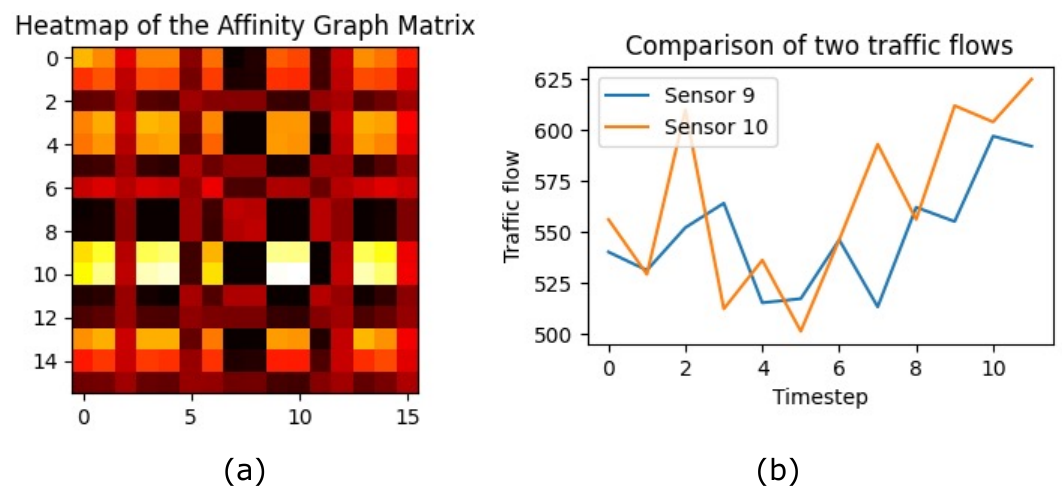}
    \caption{The case study of the proposed model. The left (a) is the affinity graph matrix, while the right (b) is the traffic flow data of two sensors (\emph{i.e.,} sensor 9 and sensor 10).}
    \label{fig:case}
\end{figure}

\subsection{Case Study}
This section provides a case study to showcase the model's ability to capture spatio-temporal dependencies and diverse transitions, and the results are shown in Figure \ref{fig:case}. The left of the figure shows the affinity graph matrix, and the right of the figure shows the traffic flow data of two sensors (\emph{i.e.,} sensor 9 and sensor 10). As can be seen from the visualization of the affinity matrix, sensor 9 and sensor 10 (indicated by the 10-th row and 9-th column of the matrix) have very high affinity scores, which shows that they are highly related in terms of traffic flow features. The right part of the figure demonstrates this, in which the traffic flow of sensor 9 and sensor 10 shows similar patterns and the traffic flow data of sensor 9 is one step behind the data of sensor 10. This shows that our model can capture spatial and temporal correlations in the traffic flow data.

\subsection{Potential Limitations and Drawbacks}
\smallskip
\noindent\textbf{Limited Exploration of Prior Information.}
While the \method{} introduces prior information into heterogeneous graphs, the effectiveness of capturing high-order spatio-temporal relationships from this information may be limited. The model's performance might be sensitive to the quality and relevance of the introduced prior knowledge, and the impact of different types of prior information on forecasting accuracy needs to be further explored.

\smallskip
\noindent\textbf{Dependency on Constructed Affinity and Penalty Graphs.}
The reliance on constructed affinity and penalty graphs to model dynamic relationships introduces an additional layer of complexity. The performance of the \method{} may be influenced by the accuracy of constructing these graphs, and the approach's robustness to variations in data quality or noise in the construction process should be investigated.

\smallskip
\noindent\textbf{Adaptability to New Traffic Scenarios.}
The effectiveness of the proposed \method{} in capturing diverse transitional patterns is demonstrated on benchmark datasets. However, its adaptability to new or unseen traffic scenarios, such as emerging traffic patterns in rapidly changing urban environments, is an open question and requires further investigation.

\section{Conclusion}

This paper studies traffic flow forecasting and proposes a novel method named \method{} to solve it. Our proposed \method{} conjointly explores high-order spatio-temporal relationships from both prior and posterior information. In particular, we not only extend road networks into heterogeneous graphs for prior message passing, but also model dynamic relationships using both affinity graphs and penalty graphs for posterior message passing. Moreover, we develop a conjoint self-attention decoder to capture diverse temporal properties in traffic data. Experimental results on three traffic datasets demonstrate the superiority of our proposed model \method{}, which outperforms the state-of-the-art baselines.

In the future, we plan to extend our GNN-based traffic forecasting model by incorporating additional data sources, such as real-time weather and events data, to further improve prediction accuracy. Additionally, we aim to explore the integration of reinforcement learning techniques to optimize traffic signal control and reduce congestion. Furthermore, investigating the scalability of our model to larger urban networks and evaluating its robustness under various traffic conditions are important directions for future research. Finally, we will continue to explore novel methods for interpretability and visualization of the model's predictions to enhance its practical utility for urban traffic management.

\section*{Acknowledgement}
This paper is partially supported by National Key Research and Development Program of China with Grant No. 2023YFC3341203, the National Natural Science Foundation of China (NSFC Grant Numbers 62306014 and 62276002) as well as the China Postdoctoral Science Foundation with Grant No. 2023M730057.

\bibliographystyle{model1-num-names}

\bibliography{ref}


\end{document}